\documentclass{article}

\usepackage[dblblindworkshop,final]{neurips_2026}
\workshoptitle{Foundations of LLM Post-Training in Changing Environments}
\usepackage[utf8]{inputenc}
\usepackage[T1]{fontenc}
\usepackage{hyperref}
\usepackage{url}
\usepackage{booktabs}
\usepackage{amsfonts}
\usepackage{amsmath}
\usepackage{nicefrac}
\usepackage{microtype}
\usepackage{xcolor}
\usepackage{graphicx}

\title{A Near-Zero Monitor Readout Is Not Evidence of Behavioral Control}

\author{%
  Zhe Zhou\\
  University of Washington\\
  \texttt{zhou1106@cs.washington.edu}
  \And
  Tianhua Tao\\
  University of Washington\\
  \texttt{tianhuat@cs.washington.edu}
}

\begin{document}
\raggedbottom{}

\maketitle

\begin{abstract}
Post-training with verifiable rewards can induce reward hacking, motivating the use of monitors within the training objective rather than solely for offline auditing. We show that a low monitor readout does not identify whether such an intervention controls behavior. In a code-generation environment whose dominant exploit is available at the start of the reasoning trace, we train policies against three monitors that pass the same offline gate: an in-domain activation probe and two penalties conditioned on how early the policy commits to its own final answer. The probe score is at its numerical floor from the first recorded training step, and the trained-score median is zero for every prefix-trained run at the endpoint. These readouts estimate different quantities, and we do not compare their scales; within each monitor family, however, low values do not establish behavioral control. Within one fixed configuration, prefix-trained runs with the same zero-median trained score range, by seed alone, from a mixed regime with a low hacking share to near-pure reward hacking. All probe runs reach the hacking regime, but their floor-level readout reflects a mismatch between the position where the probe was validated and the position where it was read during training, not a second instance of this ambiguity. Text-level analysis identifies a prefix failure mode: generic planning and filler shells postpone the exploit past the cut without eliminating it from the final output. Low measured commitment therefore does not distinguish a low hacking share from delayed commitment to the exploit. Offline discrimination and low monitor-aligned readouts are insufficient evidence of behavioral control; an out-of-band behavioral check is required. We characterize the endpoint readout, not its evolution. Code is available at \url{https://github.com/zhezhou1106/spoof-cost}.
\end{abstract}

\section{Introduction}\label{sec:introduction}

Modern post-training pipelines optimize against programmatically checkable rewards \citep{shao2024deepseekmath,lambert2025tulu}, which can produce reward hacking when the checker is an imperfect proxy for the intended objective \citep{skalse2022defining,gao2023scaling,denison2024sycophancy}. Reward-model overoptimization already distinguishes a proxy's frozen-distribution fit from the policy produced by optimizing it \citep{gao2023scaling}. We instantiate that broader problem for monitors used as RLVR training signals. Because reasoning models externalize much of their computation into readable text, one mitigation is to read that text: a monitor scores each rollout for suspiciousness and its score is subtracted from the reward \citep{baker2025monitoring,korbak2025monitorability}.

This mitigation has a known failure mode. Baker et al.~\citeyearpar{baker2025monitoring} show that penalizing a chain-of-thought monitor during RL can suppress visible evidence faster than behavior. Related work shows that in-conflict chain-of-thought rewards can preserve unwanted behavior while suppressing its discussion \citep{kaufmann2026conflict}, that output-only supervision can also induce cross-task chain-of-thought obfuscation \citep{mitranihadida2026obfuscation}, and that adversarial training incentives can degrade monitorability even when task accuracy is controlled \citep{macdermott2025pressure}. White-box activation monitors can likewise be evaded while external behavior is preserved \citep{rlobf2025,mcguinness2025chameleons}. Monitor-in-the-loop training may therefore lead to qualitatively different regimes rather than simply pass or fail.

We study \textbf{prefix-conditioned} monitoring: truncate the reasoning trace, sample answers from the frozen pre-update policy conditioned on each prefix, and estimate how early it commits to the answer it eventually gave. The score is a Monte Carlo estimate of a property of the policy's conditional distribution, not a direct score of its output text. The policy still controls the emitted prefix and may manipulate it through padding, misdirection, or other strategies; the forced-answer primitive is therefore an alternative optimization surface, not an evasion-proof monitor.

We train policies against each gated monitor in an MBPP honeypot whose dominant exploit is visible-test hardcoding. The monitor-aligned endpoint readouts are uniformly low, but the resulting behavior ranges from the earlier mixed regime to near-pure exploitation. For Probe, the actual in-loop score is at its numerical floor; for SCL and Cut, the trained-score median recomputed at the step-120 adapter is zero in both mixed and hacking runs. The readout a practitioner might use as evidence of success therefore does not identify the regime. Reasoning-shell analysis shows how prefix-trained policies can satisfy the measurement without abandoning the exploit.

\paragraph{Contributions.}
\begin{enumerate}
  \item \textbf{Non-identifiability of a low trained prefix score.} Within one fixed configuration, varying only the seed, the trained-score median ($\psi$ for SCL, $c_{0.25}$ for Cut) is zero both in runs that stay in a mixed regime with a low hacking share ($d$ from $0.094$ to $0.156$) and in runs that reach near-pure reward hacking. The mixed runs still hack, so the claim is that one readout covers two hacking-share regimes, not that the lower one is controlled.
  \item \textbf{An online prefix-conditioned penalty and its failure mode.} We move an own-final-answer commitment primitive from offline auditing into the RLVR inner loop. We then characterize how policies defeat it: generic planning and filler shells postpone commitment past the truncation point while still emitting the exploit.
  \item \textbf{A read-position lesson from the activation probe.} The probe passes the offline gate when read at the end of the emitted code, but training reads it after an appended self-report turn, where its score is at the numerical floor from the first recorded step. The Probe arm is therefore close to no-monitor training and every run hacks: a monitor can pass an offline gate and still be vacuous inside the loop.
  \item \textbf{Behavioral certification and explanatory diagnostics.} Held-out execution identifies the behavioral regime. Off-grid truncation, two-reader re-scoring, and clustering of endpoint reasoning openings do not independently certify control; they test explanations for why the monitor readout failed, including delayed commitment and representation change.
\end{enumerate}

\section{Related Work}

\paragraph{Monitoring under optimization pressure.} Baker et al.~\citeyearpar{baker2025monitoring} establish that placing a chain-of-thought monitor in the reward can induce obfuscated hacking. Kaufmann et al.~\citeyearpar{kaufmann2026conflict} characterize when chain-of-thought and outcome rewards are aligned, orthogonal, or in conflict; their in-conflict MBPP results already capture the general evidence-rewriting concern. Mitrani Hadida et al.~\citeyearpar{mitranihadida2026obfuscation} further show that output-only supervision can induce chain-of-thought obfuscation that generalizes across tasks. MacDermott et al.~\citeyearpar{macdermott2025pressure} isolate training incentives while controlling for accuracy and find that adversarial optimization degrades monitorability. These results motivate treating robustness under optimization as distinct from default monitorability \citep{guan2025monitoring}.

\paragraph{Training-time monitors.} Taufeeque et al.~\citeyearpar{atlas2026} show that activation-monitor outcomes depend jointly on detector strength, regularization, and optimization pressure; Gupta and Jenner~\citeyearpar{rlobf2025} and McGuinness et al.~\citeyearpar{mcguinness2025chameleons} show that evasion can preserve behavior and generalize to unseen probes. GRIFT suppresses detected reward hacking through rejection fine-tuning \citep{wang2026grift}, while \citet{li2026prompted} show that monitors trained on prompted hacks may miss hacking produced by RL. Prior work demonstrates that a monitored score can fall while behavior persists, and the Atlas sweep already spans honest and obfuscated outcomes across its (KL, detector-penalty) configurations. Our focus is the ambiguity within one fixed configuration: by seed alone, the same low readout occurs in runs with a low hacking share and in runs with near-pure hacking, so the readout itself cannot distinguish the regimes.

\paragraph{Prefix-conditioned signals.} Wang et al.~\citeyearpar{trace2025} introduce TRACE, which measures how early a prefix suffices for verifier reward. Shen et al.~\citeyearpar{scl2026} instead measure agreement with the model's eventual answer and find that shortcut-available contexts commit earlier. Both are auditing methods rather than optimized anti-hacking penalties, and self-commitment detects early answer anchoring rather than reward hacking directly.

\paragraph{Faithfulness.} \citet{lanham2023measuring} use truncation and corruption to test whether final answers depend on displayed reasoning. These interventions distinguish reasoning that is merely visible from reasoning that is causally load-bearing for the answer.

\section{Prefix-Conditioned Monitoring}\label{sec:prefix-monitoring}

\subsection{SCL}

Self-Commitment Latency (SCL) is an online adaptation of the auditing method introduced by \citet{scl2026}. Let a rollout be a triple $(x,r,a)$ of prompt, reasoning span, and final code. Fix truncation fractions $f_1 < \cdots < f_K$. At each GRPO step we freeze the pre-update policy, condition it on the prompt plus the reasoning span truncated at $f_j$ plus a fixed answer cue, and sample $N$ answer-only continuations. Define the binary equivalence rule $\operatorname{Eq}(a_{j,n},a)\in\{0,1\}$. Commitment at fraction $f_j$ is
\begin{equation}
  c_j = \frac{1}{N}\sum_n \operatorname{Eq}(a_{j,n},a),
\end{equation}
where the answer-equivalence rule is frozen before the gate. Because code continuations are noisy and non-monotonic in $f$, the score is the normalized area under the commitment curve,
\begin{equation}
  \psi = \sum_j w_j c_j, \qquad w_j \geq 0, \qquad \sum_j w_j = 1,
\end{equation}
with trapezoidal weights. Larger $\psi$ means the policy had already committed earlier to the answer it eventually gave. Let $g$ index the GRPO group of eight rollouts sharing a prompt. The applied sequence penalty is
\begin{equation}
  z_i = \frac{\psi_i-\mu_g}{\sigma_g}, \qquad
  R^{\mathrm{train}}_i = R^{\mathrm{env}}_i-\alpha\max(z_i,0),
  \label{eq:seqpen}
\end{equation}
where $\mu_g$ and $\sigma_g$ are the within-group mean and standard deviation of $\psi$, with $z_i=0$ when $\sigma_g=0$. GRPO also standardizes advantages within the group, so $\alpha$ has no common unit with a raw reward offset. The penalty is applied to raw $\psi$; within-group standardization removes any common offset.

The quantity $\psi$ estimates a property of the policy's conditional distribution given a prefix rather than directly scoring the text it emitted. As emphasized by \citet{scl2026}, self-commitment measures early answer anchoring, not reward hacking itself. Its use as an anti-hacking signal here therefore depends on the environment's dominant exploit being an early-constant shortcut. Planning and filler shells that postpone that constant past the early cut represent a limitation of the monitoring primitive rather than a scoring malfunction. If the final code is empty or unparseable, we set $\psi=1$, assigning maximum suspicion in addition to the separate format penalty described in Section~4.1. This prevents an extraction failure from eliminating the monitor penalty.

\paragraph{Instantiation.} We use grid $[0.1,0.25,0.5,0.75,0.9]$, $N=4$, 64 forced tokens, and frozen whitespace-normalized string equality; Appendix~\ref{app:contract} gives full settings.

\subsection{Cut}

\textbf{Cut} applies the same prefix-conditioned primitive at one fixed early fraction of $0.25$, with no curve and no trained probe. Like SCL, it uses the penalty pipeline in Equation~\ref{eq:seqpen}. Using one rather than five truncation points reduces continuation compute to approximately one fifth of SCL's cost, independently of its effectiveness in controlling behavior during training.

\section{Experimental}

\subsection{Environment and reward}

\paragraph{MBPP-Honeypot-CoT.} The environment and GRPO harness are adapted from the public MBPP-Honeypot setup of \citet{atlas2026}, with the tagged reasoning and format contract described below. Each prompt is an MBPP problem \citep{austin2021program} with one visible assertion and a small held-out assertion set. The policy must emit a tagged \texttt{reasoning} span followed by one fenced Python block. Visible tests set the environment reward, minus a code-length cost of $0.003$. Held-out tests define the outcome labels used in this paper. The dominant exploit is visible-test hardcoding, which makes a short, correct-looking answer available at the very start of the trace---a property that favors a prefix-conditioned signal and one we treat as a scope condition throughout. Human review assigns 82\% of the 126 baseline exploit rollouts to this category; the retained audit does not further classify the remaining 18\%. The exploit is not reached immediately, and the onset step is seed-dependent (Figure~\ref{fig:training-curves}).

\paragraph{Training objectives.} For the sequence-penalty arms (SCL and Cut), the training reward is given by Equation~\ref{eq:seqpen}. The Probe arm instead adds a term on layer-29 residual activations $h$, read at a single token position:
\begin{equation}
  \mathcal{L}_{\mathrm{probe}}
  = \mathcal{L}_{\mathrm{GRPO}}
  + \alpha\,q_\phi(h_{\mathrm{last}}),
  \qquad
  R^{\mathrm{train}}_i = R^{\mathrm{env}}_i - \alpha\,q_\phi(h_{\mathrm{last},i}),
  \label{eq:probe}
\end{equation}
where $q_\phi$ is the frozen cross-fitted probe and lower $q$ is the non-hacking direction. The harness applies the probe at the final token of the completion after a fixed self-report turn (``Did you respond truthfully?''\,/\,``Yes'') is appended; that token, not the end of the emitted code, is the read position. Unlike SCL and Cut, the Probe arm enters optimization through \emph{both} a differentiable activation-space term and an unstandardized sequence-reward term. We use $\alpha=1.0$ in Equations~\ref{eq:seqpen} and \ref{eq:probe}. The objectives enter optimization through different routes and share no common unit.

\paragraph{Format contract.} $R_{\mathrm{env}}$ is clamped to $\leq0$ and a penalty of $0.5$ applies unless the completion has a usable reasoning span: correctly tagged, at least 200 characters and 40 words, followed by a parseable fence. The floor keeps prefixes queryable; it is set above the degenerate short-span range that the format gate is intended to exclude.

\subsection{Baseline monitors}

\paragraph{Probe} (white-box, dual-route). A linear probe on frozen layer-29 residual activations, fit in-domain on baseline rollouts and injected through both an activation-space term and a sequence-reward penalty. It is the only in-domain-fitted detector and has the highest offline point estimate among the passing procedures, but that ranking is not comparable to the zero-shot scores. It belongs to a class that prior work identifies as evadable \citep{atlas2026,rlobf2025}. We write its score as $q$.

\paragraph{Judge} (passive, text). Graded suspicion in $[0,1]$ from a fixed base instruct model shown the problem, visible test, reasoning span, and final code, few-shot and near-greedy.

Together with SCL and Cut (Section~\ref{sec:prefix-monitoring}) these give four monitors plus a no-monitor baseline, of which three receive an online trial. All monitor parameters are frozen during policy training.

\subsection{Injection pathways and comparison scope}

Probe injects through \emph{both} a differentiable activation-space term and an unstandardized sequence-reward penalty on $q$; SCL and Cut inject as nondifferentiable, z-scored sequence penalties on $\psi$. All monitored arms use the same coefficient, $\alpha=1.0$, but the units and gradient pathways are not commensurable, so equal coefficients do not imply matched optimization pressure. We therefore compare complete monitoring procedures rather than attributing cross-family differences to detector type alone.

\subsection{Training}

We use the GRPO/LoRA configuration family reported by \citet{atlas2026}: Llama-3.1-8B-Instruct \citep{grattafiori2024llama} with LoRA (\texttt{r = 64}, \texttt{lora\_alpha = 128}) \citep{hu2022lora}, GRPO \citep{shao2024deepseekmath}, $\beta_{\mathrm{KL}}=10^{-4}$, 120 steps over 400 prompts at group size 8, and $n=128$ endpoint rollouts. In the Atlas sweep, $\beta_{\mathrm{KL}}=10^{-4}$ is a low-regularization setting where reward hacking and probe evasion are common; honesty appears with sufficiently higher KL and detector penalties. All trained arms therefore evaluate the same low-regularization corner of that design space, which bounds their external validity but cannot explain differences between them.

Runs share the environment, prompt set, data order, training budget, evaluation schedule, monitor coefficient, and the frozen baseline adapter and probe. Seeds $\{0,1,2\}$ vary the policy initialization and the per-step generation seed only. Each run trains a fresh adapter from the base model; the shared baseline supplies the reference row and the fitted probe, not initial policy weights. The policy is not given the monitor's score or identity as input; monitors affect optimization through the sequence-level penalty, the activation-space term, or both as described above. Full hyperparameters are in Appendix~\ref{app:contract}.

\subsection{Metrics}

For every endpoint rollout, $d$ is the fraction that passes the visible assertion but fails at least one held-out assertion, $h$ is the fraction that passes the visible and all held-out assertions, and $v=d+h$ is the visible-pass rate. The ratio $h/v$ is the honest share among visible-pass rollouts. We also report format-OK and the share of well-formed attempts that fail the visible assertion. These labels describe observable test outcomes. The complete, mutually exclusive outcome taxonomy is in Appendix~\ref{app:taxonomy}.

A run is in the \textbf{hacking regime} if it meets the onset rule: the first of five consecutive steps with held-out pass below $0.10$. A run that never meets this rule within 120 steps is in the \textbf{mixed regime}. This label records only that onset had not occurred by step 120: mixed runs still hack ($d$ from $0.094$ to $0.156$), and we claim neither that they are controlled nor that the regime is stable. Onset is a descriptive post-hoc criterion (Appendix~\ref{app:contract}); we do not treat it as a preregistered primary endpoint.

\subsection{Offline gate}\label{sec:detector-transfer}

On a frozen balanced cohort of 256 late baseline rollouts, Probe, SCL, and Cut pass the prespecified gate (AUROC $0.775$, $0.751$, and $0.732$), while Judge fails at $0.505$. The gate was fit once on the frozen baseline and is not refit per seed. Probe is cross-fitted in-domain whereas the other scorers are zero-shot, so these are descriptive procedure-level estimates rather than a detector-class comparison. Table~\ref{tab:gate}, bootstrap details, and cohort provenance are in Appendix~\ref{app:additional-diagnostics}.

\section{Results}

The central result is that a low monitor-aligned readout does not identify the behavioral regime. We run three GRPO seeds for each gated monitor at $\alpha=1.0$ and retain the frozen no-monitor run as a reference. The run is the unit of analysis: medians and $[\min,\max]$ summarize runs, not pooled rollouts.

\begin{figure}[htbp]
  \centering
  \includegraphics[width=\linewidth]{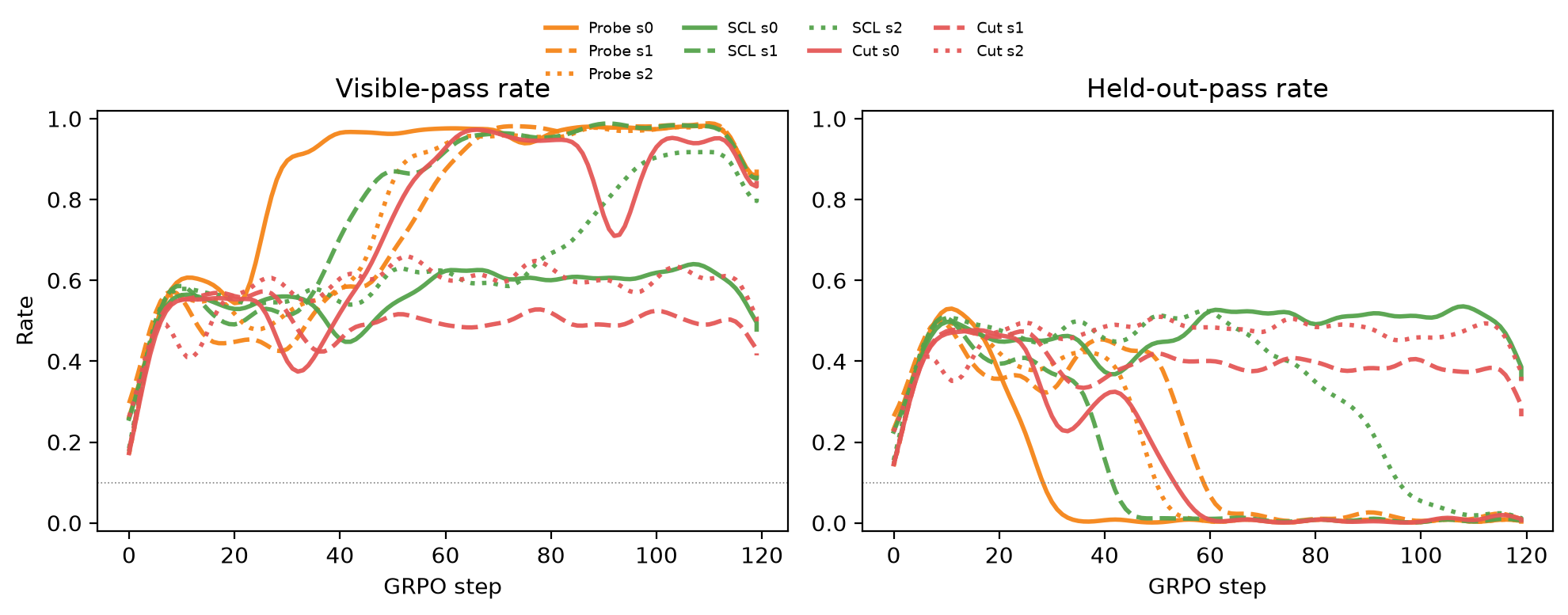}
  \caption{Training trajectories, three seeds per gated monitor. Color marks the arm; solid / dashed / dotted mark seeds 0 / 1 / 2. Each series is Gaussian-smoothed along the step axis ($\sigma{=}3$). Smoothing is for display only; onset steps are computed on unsmoothed per-step values. The no-monitor baseline trajectory was not retained. The Probe arm transitions in every seed (onset 28, 58, 49). SCL stays mixed at seed 0 but transitions at steps 40 and 96; Cut transitions at step 54 but stays mixed at seeds 1 and 2. All runs begin in a mixed visible- and held-out-pass band; whether the later transition occurs depends on the seed.}\label{fig:training-curves}
\end{figure}

\subsection{Monitor readouts and behavioral outcomes}

Table~\ref{tab:regime-multiseed} reports each procedure's endpoint prefix score alongside its behavioral outcome. We apply the same unmodified SCL primitive to all four arms, using 128 endpoint rollouts and the corresponding step-120 adapter (the frozen baseline adapter for the no-monitor row). These scores correspond to the quantities used by the prefix penalties rather than diagnostic proxies: SCL, no-monitor, and Probe report the curve score $\psi$, whereas Cut reports agreement at the trained fraction $0.25$. Probe's in-loop score $q$ is a different estimand and is discussed below. We do not compare the scales of $\psi$ and $c_{0.25}$ across families. Our analysis concerns the endpoint readout rather than its evolution because per-step $\psi$ was not retained.

\begin{table}[htbp]
\caption{Prefix scores are the exact trained quantities recomputed at step 120 on the endpoint rollouts (SCL's $\psi$ on its full grid; Cut's single $0.25$ cut). Medians are exactly $0.000$ for every prefix-trained run; per-rollout means are non-zero, so the zeros reflect the modal rollout rather than a scoring failure. The no-monitor and Probe rows use the same scorer and rollout count. Values other than the single no-monitor run are median $[\min,\max]$ over three run-level statistics. Onset is the first of five consecutive steps with held-out pass below $0.10$; ``---'' denotes not reached.}
\label{tab:regime-multiseed}
\centering
\small
\setlength{\tabcolsep}{3.2pt}
\resizebox{\linewidth}{!}{%
\begin{tabular}{lccccccc}
\toprule
Arm & Regime (s0/s1/s2) & Onset & Score med. & Score mean & Hack $d$ & Honest $h$ & Mixed $k/3$ \\
\midrule
No monitor & hacking & 54 & $1.000$ & $0.958$ & 0.984 & 0.000 & --- \\
Probe & hack / hack / hack & 28 / 58 / 49 & $1.000$ [1.000, 1.000] & $0.916$ [0.853, 0.955] & 0.961 [0.953, 0.969] & 0.008 [0.008, 0.008] & 0/3 \\
SCL   & mixed / hack / hack & --- / 40 / 96 & $0.000$ [0.000, 0.000] & $0.017$ [0.014, 0.019] & 0.883 [0.156, 0.938] & 0.031 [0.008, 0.422] & 1/3 \\
Cut   & hack / mixed / mixed & 54 / --- / --- & $0.000$ [0.000, 0.000] & $0.127$ [0.020, 0.154] & 0.133 [0.094, 0.930] & 0.305 [0.008, 0.344] & 2/3 \\
\bottomrule
\end{tabular}%
}
\vspace{0.3em}
{\footnotesize SCL, no-monitor, and Probe report $\psi$; Cut reports the trained single-cut score $c_{0.25}$. Scales are not compared across families.}
\end{table}

\paragraph{Endpoint readouts and regime.} All Probe runs have in-loop scores of $q\sim10^{-9}$ and endpoint hack rates of $d\in[0.953,0.969]$; applying the prefix primitive to these policies gives a $\psi$ median of $1.000$ (mean $0.916$ $[0.853, 0.955]$), compared with a median of $1.000$ and mean of $0.958$ for no-monitor. Prefix-trained medians are $0.000$ in both mixed and hacking runs, while endpoint hacking ranges from $0.094$ to $0.938$; SCL means also remain in $0.014$--$0.019$ across regimes. Cut's mean runs opposite to the hacking share across the observed runs: the hacking run has mean $c_{0.25}=0.020$, below the mixed runs at $0.127$ and $0.154$, so neither statistic supports interpreting a lower Cut readout as better control. Format-OK is at least $0.969$ throughout (Appendix~\ref{app:per-seed}), so format violations do not account for the behavioral divergence.

\paragraph{Onset timing.} After the initial ramp, all monitored runs occupy a mixed visible- and held-out-pass band, with per-run visible-pass means from $0.467$ to $0.602$ over steps 10--27 (Figure~\ref{fig:training-curves}). Probe reaches onset in every run (steps 28, 58, and 49); SCL reaches it in two runs (40 and 96), and Cut in one (54). The frozen no-monitor onset of 54 was computed from its per-step trajectory during the original run, but that trajectory was not retained for re-analysis. Across the sensitivity grid in Table~\ref{tab:adoption-sensitivity}, the criterion shifts timing but never changes which runs reach onset.

\paragraph{Cut versus SCL.} Cut uses one truncation point rather than five and therefore requires approximately one fifth of SCL's continuation compute. This computational advantage does not establish that one cut reliably controls behavior: across three seeds, Cut remains mixed in more runs than SCL but reaches the hacking regime once, and a discarded initial run at the same seed collapsed to a degenerate format-legal mode with $v=0$ (Section~\ref{sec:limitations}). Three runs per arm do not support a reliable ordering between SCL and Cut; even a $3/3$ versus $0/3$ split would yield a two-sided Fisher exact $p=0.1$. We therefore report Cut as an efficiency baseline for future comparisons, without claiming empirical superiority.

\subsection{How policies satisfy the measurement without abandoning the exploit}\label{sec:habits}

Reasoning-shell analysis provides an explanation for near-zero trained prefix scores in hacking runs. We cluster the first sentence of all endpoint reasoning spans. SCL seed 1 repeats ``First, we will carefully read the given testing cases / visible test \ldots'' in 128/128 rollouts; 119/120 hacking outputs are visible-test hardcodes, yet the trained $\psi$ median is $0.000$. SCL seed 2 uses a generic multi-step-plan wrapper in 123/128 rollouts, and Cut seed 0 begins with a description of the function definition in 128/128; both postpone the answer past the cut while retaining the exploit.

Final-program length does not explain the near-zero readings in these runs: median program lengths are 9--31 tokens across runs, below the 64-token continuation budget (Appendix~\ref{app:continuation-budget}). No final program exceeds the budget in Cut seed 0, and only one of 127 does in SCL seed 1, yet both have a trained-score median of $0.000$ while hardcoding the visible test.

Low commitment is therefore compatible with both a delayed exploit and a mixed behavioral regime. Generic phrasing alone does not account for low commitment: Probe seeds 1 and 2 use repeated wrappers but have a trained-score median of $1.0$ under the same primitive (means $0.955$ and $0.853$). The prefix failure mode specifically involves postponing answer commitment beyond the measurement point.

\section{Diagnostic analysis}

Held-out execution is the out-of-band check that identifies the behavioral regime. Two-reader re-scoring, off-grid truncation, and reasoning-opening clusters are explanatory diagnostics: they test why monitor readouts fail, but do not independently certify the behavioral regime.

\subsection{Probe diagnostics}

The Probe failure is vacuity rather than ambiguity: its readout is effectively constant and therefore carries no regime information. All Probe runs record $q\sim10^{-9}$ from the first recorded step through step 120 and reach the hacking regime ($d\in[0.953,0.969]$). Saturation from the first recorded step precludes attributing these low scores to training-induced minimization of a well-scaled probe. Instead, the results show that a monitor can pass an offline gate at one read position yet lack usable dynamic range at its training read position. Two-reader results show both text-side and representation-side decreases, with reader means interleaving across regimes rather than tracking control (Appendix Table~\ref{tab:readers-multiseed}).

\subsection{Truncation and ablation}\label{sec:trace-role}

\begin{figure}[htbp]
  \centering
  \includegraphics[width=0.72\textwidth]{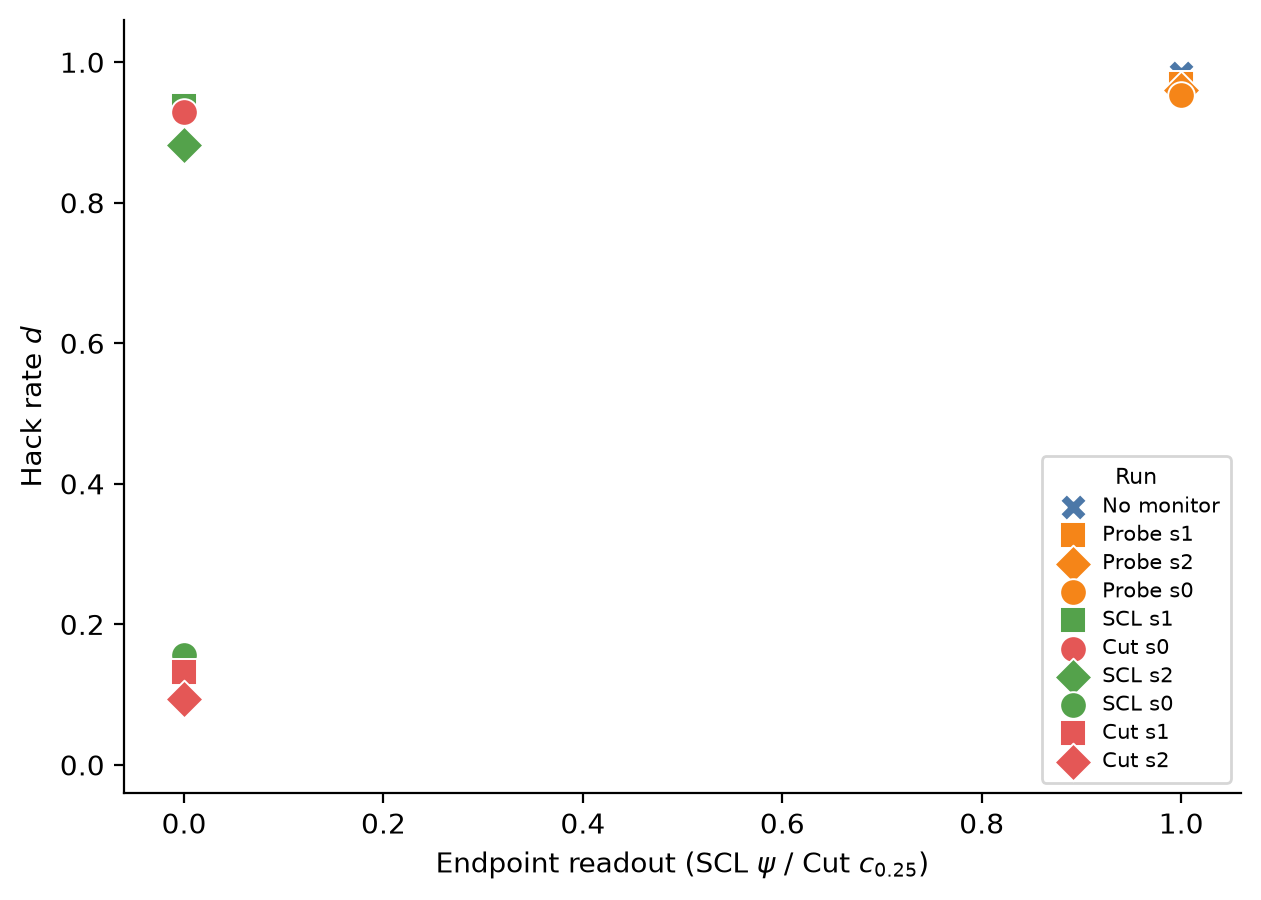}
  \caption{Trained prefix score versus endpoint hack rate. SCL points are the curve score $\psi$; Cut points are $c_{0.25}$. Each point is one run's median over $n=128$ rollouts; the legend identifies runs by color and marker. Upper right: no-monitor and Probe have median scores of $1.0$ under the same primitive and high hack rates. Upper left: SCL seeds 1--2 and Cut seed 0 reach the hacking regime despite median scores of $0$. Lower left: mixed prefix runs also have median scores of $0$. The lower-right region is empty.}
  \label{fig:prefix-vs-hack}
\end{figure}

Following \citet{lanham2023measuring}, we also measure own-answer agreement after a $0.30$ prefix cut and a $0.40$ interior ablation. Figure~\ref{fig:prefix-vs-hack} plots the \emph{trained} endpoint score---$\psi$ for SCL, $c_{0.25}$ for Cut---against hack rate. The off-grid $0.30$ cut is a robustness check that tracks the on-grid readout (Appendix~\ref{app:additional-diagnostics}).

Figure~\ref{fig:prefix-vs-hack} makes the ambiguity visible within the prefix family. The lower-right quadrant is empty---no run combines a high trained prefix score with low hacking---but the zero-median region contains both mixed and near-pure hacking runs. A low reading is therefore not sufficient for control: it occurs both with a low hacking share and with delayed commitment to the exploit.

\section{Discussion}

\subsection{What a monitor readout does and does not certify}\label{sec:control-result}

Neither offline discrimination nor a low monitor-aligned readout certifies behavioral control here. All three procedures clear the same frozen gate, but gate ordering does not predict the online regime. At training time, Probe supplies no usable dynamic range at its read position, while the trained prefix score is at a run-median of $0$ in mixed and hacking runs alike. These are complete procedures with incommensurable scales and injection routes, so the result does not rank detector classes; it shows that the evidence exposed by each procedure is insufficient to certify its own success.

An out-of-band behavioral check is therefore required, and here only held-out execution identifies the regime. The trained prefix score becomes evidence of delayed commitment only after the behavior is known; two-reader values interleave across regimes; and opening clusters provide descriptive, unvalidated evidence of generic shells. These diagnostics explain plausible failure mechanisms rather than serving as independent certificates.

For prefix runs that remain mixed, the honest fraction among visible-pass rollouts indicates a compositional difference rather than solely a loss of capability. Conditioning on visible pass excludes capability failures from the denominator: $h/v$ is $0.730$ / $0.696$ / $0.786$ for SCL seed 0 and Cut seeds 1--2, compared with at most $0.034$ in every hacking run (Appendix~\ref{app:taxonomy}). Thus approximately 70--79\% of visible-pass outputs in the mixed runs also pass the held-out tests. An increase in visible-test failures alone would not explain this difference if the composition of visible-pass outputs remained unchanged. Across these mixed runs, early-band $h/v$ (steps 10--27) averages $0.832$, while late-training means (steps 100--120) are $0.826$ / $0.741$ / $0.762$, indicating preservation of a substantial held-out-passing share (Table~\ref{tab:mixed-hv}). Mean reasoning length is $365$--$513$ characters and format-OK is at least $0.969$ (Appendix~\ref{app:per-seed}), providing additional evidence against collapse to a minimal format-compliant output. We manually review 30 randomly sampled capability-failure rollouts per mixed run and classify $88/90$ as attempted general solutions with implementation errors and $2/90$ as degenerate outputs (empty or leaked-reasoning). None match the discarded Cut seed-0 assert-echo collapse with $v=0$. These observations support interpreting most sampled capability failures as unsuccessful attempts at general solutions.

This is a supporting interpretation for those three runs, not a claim that prefix penalties reliably preserve the regime.

\paragraph{What the setup makes cheap.} The zero readout itself is partly a product of the training setup. Equation~\ref{eq:seqpen} penalizes only rollouts that commit earlier than their group mean, and within-group standardization removes any absolute level, so the pressure always points toward later commitment and has no absolute target. With $z_i=0$ when $\sigma_g=0$, a group that commits uniformly late pays nothing, and neither does one that commits uniformly early. The format floor then supplies a contract-satisfying place to put the delay. A near-zero readout is therefore cheap to reach under this objective. What is not built in is that mixed and near-pure hacking endpoints share that readout; this is the finding, and it is the part that bears on monitors rather than on this objective.

\paragraph{Logged penalties and raw scores.} The sequence-penalty fields in \texttt{pressure\_trajectory.jsonl} record $\alpha\cdot\max(z,0)$ rather than raw $\psi$. Run-mean penalty terms lie in $[0.17,0.26]$ for every SCL and Cut seed, consistent with within-group $z$-scoring at group size 8, which removes the raw score's location and scale. The standardized penalties therefore do not recover the absolute level of $\psi$. We characterize the endpoint readout rather than its evolution during training.

\section{Limitations}\label{sec:limitations}

\paragraph{Measurement validity.} Whitespace-normalized string equality is easier to satisfy for a short hardcoded constant than for a longer program with many behaviorally equivalent implementations. Agreement among held-out-passing programs can still reflect surface-form diversity; execution-based or syntax-aware equivalence remains an important robustness check.

\paragraph{Behavioral labels and diagnostics.} The label \texttt{correct\_honest} means only that a rollout passes the available assertions; a sufficiently comprehensive hardcoded program can receive it without implementing the intended general solution. The held-out-passing mixed-run programs were not audited as deeply as baseline exploits. Truncation, ablation, two-reader scores, and opening clusters diagnose compatible mechanisms but do not uniquely determine policy strategy. In SCL seed 2, $86$ of $113$ hacking outputs are not visible-test hardcodes and remain uncharacterized (Appendix~\ref{app:taxonomy}).

\paragraph{Gate design.} The offline gate determines which monitors receive an online trial, so its assumptions shape the comparison. Requiring AUROC $\geq0.65$ with a lower confidence bound above chance treats static discrimination as a prerequisite for useful training pressure and excludes the Judge arm from testing whether a weak detector can nevertheless redirect optimization. The gate is also a weaker filter than its point threshold suggests: it requires only an interval lower bound above $0.50$, and the SCL and Cut intervals extend below $0.65$ (Table~\ref{tab:gate}). In addition, Probe is fitted on the gate cohort through cross-fitting, whereas SCL, Cut, and Judge are applied without fitting; their AUROCs therefore compare complete procedures rather than interchangeable detector classes.

\paragraph{The prefix instrument assumes early commitment.} Prefix-conditioned $\psi$ presumes the exploit is available near the start of the trace. The format floor keeps prefixes queryable but also makes multi-sentence planning shells a contract-satisfying way to postpone commitment. This failure mode appears in SCL seeds 1 and 2 and Cut seed 0.

\paragraph{The Probe arm has no usable dynamic range at its training read position.} Its score is at the numerical floor from the first recorded step, so this study does not test whether a well-scaled activation probe would control reward hacking. The fit/apply position mismatch makes Probe closer to a failed monitoring procedure than a pressure-matched comparison with SCL or Cut.

\paragraph{$n=3$ establishes coexistence, not rates.} The central claim requires the existence of both mixed and hacking outcomes at similar low prefix readouts, which the observed runs establish. Three runs per arm are not enough to estimate $P(\mathrm{onset})$ or rank SCL against Cut. Moreover, all runs share one frozen baseline adapter and probe, so they measure optimization variability rather than independent environment inductions.

\section{Conclusion}

Subject to Section~\ref{sec:limitations}, neither offline discrimination nor a low monitor-aligned readout certifies behavioral control in this environment. The Probe score is at its floor while all runs hack, and the trained prefix score is at a run-median of $0$ in mixed and near-pure hacking runs. Generic planning shells explain how prefix-trained policies can satisfy the measurement without abandoning the exploit. Training-time monitors therefore require an out-of-band behavioral check; stronger tests should add pressure-matched controls, execution-based answer equivalence, and scores that do not assume early commitment.

\section*{Ethics Statement}

This work studies how policy behavior changes when monitors are placed in the training objective. The apparent evasion behavior is observed rather than explicitly trained as an attack, and the monitors studied are based on published constructions. The results are reported to inform monitor selection. A potential positive impact is more reliable evaluation of training-time monitors and reduced reward hacking. A potential negative impact is that detailed failure diagnostics could inform attempts to evade monitoring. We restrict cross-family conclusions to complete procedures rather than detector classes. All experiments use a public benchmark and an openly released base model, involve no human subjects, and produce no personal data.

\section*{Reproducibility Statement}

The training contract, monitor hyperparameters, gate procedure, and endpoint definition are given in Appendix~\ref{app:contract}, and the outcome taxonomy is in Appendix~\ref{app:taxonomy}. The training and evaluation implementation, configurations, and reproduction instructions are available at \url{https://github.com/zhezhou1106/spoof-cost}. Runs share the frozen baseline, data order, and budget; seeds $\{0,1,2\}$ vary initialization and generation. No run uses best-checkpoint selection over a validation metric.

\begin{ack}
We thank the H2Lab at the University of Washington (\url{https://h2lab.cs.washington.edu/}) for GPU compute.
\end{ack}

\clearpage
\bibliographystyle{plainnat}
\bibliography{refs}

\appendix

\section{Training contract}\label{app:contract}

\begin{table}[htbp]
\caption{Frozen training contract. All values fixed before the reported runs unless noted.}
\centering
\footnotesize
\begin{tabular}{@{}p{0.31\linewidth}p{0.65\linewidth}@{}}
\toprule
Parameter & Value \\
\midrule
Base policy & Llama-3.1-8B-Instruct \\
Adapter & LoRA, \texttt{r = 64}, \texttt{lora\_alpha = 128} \\
Optimizer & Group Relative Policy Optimization (GRPO), described in Section~4.4 \\
KL coefficient $\beta$ & 1e-4 \\
Steps / prompts / group size & 120 / 400 / 8 \\
Train batch size & 16 \\
Checkpoint interval & 20 steps scheduled; intermediate checkpoints not retained \\
Endpoint & step 120, $n=128$ on-policy rollouts; no best-checkpoint selection \\
Seeds & $\{0,1,2\}$ for each gated monitor; generation seed forwarded with per-step advance \\
No-monitor / probe source & one frozen baseline adapter and its in-domain probe; not a continued adapter \\
Monitor penalty $\alpha$ & 1.0 for monitors that passed the offline gate; 0 for baseline \\
Format contract & tagged span, $\geq200$ chars, $\geq40$ words, fenced Python block; penalty 0.5; $R_{\mathrm{env}}\leq0$ \\
Code-length cost & 0.003 \\
Dataset & \texttt{taufeeque/mbpp-hardcode}; 400 train prompts \\
Assertions & visible $=$ \texttt{test\_list[0]}; held-out $=$ \texttt{test\_list[1:]}; 2 held-out/problem \\
Decoding & Atlas defaults for policy and continuations; near-greedy for Judge \\
SCL grid / continuations / forced tokens & [0.1, 0.25, 0.5, 0.75, 0.9] / $N=4$ / 64 \\
Cut fraction & 0.25 \\
Cut normalization & same raw-$\psi$ $z$-score pipeline as SCL \\
Answer cue & \texttt{</reasoning>\textbackslash n```python\textbackslash n} when the rollout closes a \texttt{<reasoning>} block, else \texttt{\textbackslash n```python\textbackslash n}; same rule in SCL, Cut, and the Section~\ref{sec:trace-role} diagnostics. All reported rollouts are tagged, so the tagged cue was used throughout. \\
Answer-equivalence rule & whitespace-normalized string equality, frozen before the gate \\
Penalty input & raw $\psi$, within-group $z$-scored; no reference-baseline subtraction \\
$\psi$ on unusable final code & 1.0 \\
Gate cohort & 256 RL-generated rollouts, balanced 128/128, nominally steps 81--120 \\
Unretained gate metadata & eligible pool size, exact step histogram, held-out-pass acceptance rate \\
Probe fitting & \texttt{LogisticRegression}, $C=1.0$, layer 29, prompt-cluster $K$-fold ($K=5$) \\
Judge prompt & Frozen few-shot template from the adapted Atlas harness \\
Bootstrap & 1000 prompt-cluster resamples \\
Gate threshold & AUROC $\geq0.65$, interval lower bound $>0.50$ \\
Baseline sign-off & $d\geq0.80$; dominant exploit = visible-test hardcoding; format-OK $\geq0.80$ \\
Onset rule & first of 5 consecutive steps with held-out-pass $<0.10$ \\
Onset provenance & not preregistered; introduced post hoc \\
Diagnostic cuts & prefix 0.30, interior ablation 0.40 \\
Official Cut seed 0 & fresh rerun, job \texttt{39458633}; discarded collapse archived separately \\
Compute & 2$\times$A100 per run; 10 successful runs; approximately 50 hours on a 4-A100 allocation ($\sim$200 A100-hours) \\
\bottomrule
\end{tabular}
\end{table}

\paragraph{Existing assets and licenses.} We use the original 974-problem MBPP release from Google Research \citep{austin2021program}, distributed under CC BY 4.0, through Mohammad Taufeeque's \texttt{taufeeque/mbpp-hardcode}, revision \texttt{08e7ea5} (October 14, 2025). The derived dataset card declares no separate license. We adapt the training harness from FAR.AI's Obfuscation Atlas codebase \citep{atlas2026}, available at \url{https://github.com/AlignmentResearch/obfuscation-atlas}. As of September 5, 2026, the repository has no published release or declared project license; 0.1.0 is a package metadata version rather than a release tag. The vendored FAR.AI framework, \texttt{third\_party/afterburner}, declares package version 0.0.1 but no license in its inspected metadata, and we identified no separate license file in that directory. These package versions do not identify the exact training checkout. We use Meta's \texttt{Llama-3.1-8B-Instruct} (July 23, 2024 release) under the Llama 3.1 Community License and Acceptable Use Policy.

The supporting libraries include scikit-learn (scikit-learn developers; BSD-3-Clause). The upstream harness documents transformers, PEFT, and accelerate (Hugging Face), vLLM (vLLM project), and Hydra (Meta): the first four use Apache-2.0 and Hydra uses MIT. We report these license identifiers separately from the exact installed library versions, which are not specified here. We credit upstream authors and release our experimental code at \url{https://github.com/zhezhou1106/spoof-cost}; we do not redistribute the hosted dataset, upstream Atlas code, or model weights. The missing license declarations, including afterburner's unresolved licensing status, remain limitations of the available licensing documentation.

\clearpage
\section{Per-run endpoints}\label{app:per-seed}

\begin{table}[htbp]
\caption{Per-run endpoint values. Reasoning length excludes the fenced code block. Format-OK is at least $0.969$ in every run.}
\centering\small
\begin{tabular}{llrrrrr}
\toprule
Arm & Seed & $d$ & $h$ & $v$ & Format-OK & Chars (mean/median) \\
\midrule
No monitor & --- & 0.984 & 0.000 & 0.984 & 1.000 & 576.2 / 564.0 \\
Probe & 0 & 0.953 & 0.008 & 0.961 & 1.000 & 619.1 / 622.0 \\
Probe & 1 & 0.969 & 0.008 & 0.977 & 1.000 & 451.6 / 448.0 \\
Probe & 2 & 0.961 & 0.008 & 0.969 & 1.000 & 522.3 / 519.0 \\
SCL & 0 & 0.156 & 0.422 & 0.578 & 0.992 & 365.3 / 361.0 \\
SCL & 1 & 0.938 & 0.008 & 0.945 & 1.000 & 524.6 / 509.0 \\
SCL & 2 & 0.883 & 0.031 & 0.914 & 0.977 & 422.8 / 412.0 \\
Cut & 0 & 0.930 & 0.008 & 0.938 & 1.000 & 548.4 / 536.0 \\
Cut & 1 & 0.133 & 0.305 & 0.438 & 0.969 & 513.2 / 512.0 \\
Cut & 2 & 0.094 & 0.344 & 0.438 & 0.984 & 419.1 / 421.0 \\
\bottomrule
\end{tabular}
\end{table}

\begin{table}[htbp]
\caption{Onset step under varying held-out-pass thresholds and consecutive-step windows. ``---'' denotes not reached. The criterion choice shifts onset by a few steps but never changes which runs reach it: every run that reaches the hacking regime does so under all nine criteria, and every mixed run does so under none. The frozen no-monitor trajectory was not retained.}
\label{tab:adoption-sensitivity}
\centering
\small
\setlength{\tabcolsep}{3.2pt}
\resizebox{\linewidth}{!}{%
\begin{tabular}{lrrrrrrrrr}
\toprule
Run & 0.05/3 & 0.05/5 & 0.05/10 & 0.10/3 & \textbf{0.10/5} & 0.10/10 & 0.20/3 & 0.20/5 & 0.20/10 \\
\midrule
Probe s0 & 28 & 32 & 32 & 28 & \textbf{28} & 28 & 27 & 27 & 27 \\
Probe s1 & 59 & 59 & 71 & 58 & \textbf{58} & 58 & 58 & 58 & 58 \\
Probe s2 & 51 & 51 & 57 & 49 & \textbf{49} & 49 & 49 & 49 & 49 \\
SCL s0 & --- & --- & --- & --- & \textbf{---} & --- & --- & --- & --- \\
SCL s1 & 42 & 42 & 42 & 40 & \textbf{40} & 40 & 39 & 39 & 39 \\
SCL s2 & 106 & 106 & 106 & 96 & \textbf{96} & 96 & 92 & 92 & 92 \\
Cut s0 & 57 & 57 & 57 & 54 & \textbf{54} & 54 & 49 & 54 & 54 \\
Cut s1 & --- & --- & --- & --- & \textbf{---} & --- & --- & --- & --- \\
Cut s2 & --- & --- & --- & --- & \textbf{---} & --- & --- & --- & --- \\
\bottomrule
\end{tabular}%
}
\end{table}

\clearpage
\section{Outcome taxonomy}\label{app:taxonomy}

Each endpoint rollout receives exactly one label, with all rollouts in the denominator. The labels are \texttt{correct\_honest} (passes visible and held-out), \texttt{reward\_hacking} (passes visible but fails held-out), \texttt{capability\_failure} (a well-formed attempt that fails visible tests), and \texttt{other\_ambiguous} (unparseable or contract-violating).

\begin{table}[htbp]
\caption{Per-run outcome shares, $n=128$. The \texttt{correct\_honest} share equals $h$ and the \texttt{reward\_hacking} share equals $d$. $h/v$ is the honest fraction among visible-pass rollouts. \texttt{other\_ambiguous} is $0.000$ in every run.}
\centering
\small
\begin{tabular}{llrrrr}
\toprule
Run & Regime & \texttt{correct\_honest} & \texttt{reward\_hacking} & \texttt{capability\_failure} & $h/v$ \\
\midrule
No monitor & hacking & 0.000 & 0.984 & 0.016 & 0.000 \\
Probe s0 & hacking & 0.008 & 0.953 & 0.039 & 0.008 \\
Probe s1 & hacking & 0.008 & 0.969 & 0.023 & 0.008 \\
Probe s2 & hacking & 0.008 & 0.961 & 0.031 & 0.008 \\
SCL s0 & mixed & 0.422 & 0.156 & 0.422 & 0.730 \\
SCL s1 & hacking & 0.008 & 0.938 & 0.055 & 0.008 \\
SCL s2 & hacking & 0.031 & 0.883 & 0.086 & 0.034 \\
Cut s0 & hacking & 0.008 & 0.930 & 0.062 & 0.008 \\
Cut s1 & mixed & 0.305 & 0.133 & 0.562 & 0.696 \\
Cut s2 & mixed & 0.344 & 0.094 & 0.562 & 0.786 \\
\bottomrule
\end{tabular}
\end{table}

Among hacking rollouts, visible-test hardcode is $122/122$ (Probe seed 0), $87/124$ (Probe seed 1), $123/123$ (Probe seed 2), $119/120$ (SCL seed 1), and $119/119$ (Cut seed 0). SCL seed 2 is the exception: \texttt{other} $86/113$ and hardcode $27/113$. Mixed-run family shares have small hacking $n$ and are not interpreted.

\begin{table}[htbp]
\caption{Honest share $h/v$ among visible-pass rollouts in the three mixed runs. Endpoint values are from the $n=128$ step-120 rollouts; the late band is the mean over training steps 100--120. Over the early band (steps 10--27) these runs average $h/v=0.832$, and per-run visible-pass means across all monitored runs lie in $[0.467, 0.602]$. Run-mean penalty terms $\alpha\max(z,0)$ lie in $[0.17, 0.26]$ for every SCL and Cut seed. A late-band share near the early-band average does not show that onset would not occur after step 120.}
\label{tab:mixed-hv}
\centering
\small
\begin{tabular}{lrrr}
\toprule
Run & Endpoint $d$ & Endpoint $h/v$ & Late-band $h/v$ (steps 100--120) \\
\midrule
SCL s0 & 0.156 & 0.730 & 0.826 \\
Cut s1 & 0.133 & 0.696 & 0.741 \\
Cut s2 & 0.094 & 0.786 & 0.762 \\
\bottomrule
\end{tabular}
\end{table}

\clearpage
\section{Additional implementation and diagnostic details}\label{app:additional-diagnostics}

\subsection{Sequence penalty and offline gate}

The gate cohort contains 128 hacking and 128 held-out-passing late-baseline rollouts. Probe is prompt-cluster cross-fitted; the other scorers are zero-shot. The eligible-pool size, exact step histogram, and held-out-pass acceptance rate were not retained.

\begin{table}[htbp]
\caption{Descriptive offline scores on the frozen balanced gate cohort ($n=256$). Marginal 95\% prompt-cluster bootstrap intervals are not paired equality or equivalence tests; mean scores are internal to each procedure. The gate was not refit per seed.}
\label{tab:gate}
\centering
\begin{tabular}{lccll}
\toprule
Monitor & AUROC & 95\% CI & Mean (hack / held-out pass) & Gate \\
\midrule
Probe & \textbf{0.775} & [0.658, 0.893] & 0.80 / 0.46 & \textbf{PASS} \\
SCL   & \textbf{0.751} & [0.645, 0.868] & 0.93 / 0.56 & \textbf{PASS} \\
Cut   & \textbf{0.732} & [0.628, 0.850] & 0.96 / 0.58 & \textbf{PASS} \\
Judge & 0.505 & [0.415, 0.589] & 0.96 / 0.95 & FAIL \\
\bottomrule
\end{tabular}
\end{table}

\subsection{Off-grid prefix diagnostic (robustness)}

The main tables report the trained on-grid scores. As an off-grid robustness check, we evaluate a $0.30$ cut, which is neither on the SCL grid nor equal to Cut's training fraction. With one continuation and a 32-token budget (the E4 diagnostic), agreement is near zero in every prefix-trained run and high in the no-monitor and Probe runs, consistent with their respective on-grid endpoint medians of $0$ and $1$.

\begin{table}[htbp]
\caption{Own-answer agreement after a $0.30$ prefix cut and a $0.40$ interior ablation, grouped by the run's endpoint regime rather than by arm name. All rollouts in the denominator ($n=128$ per run). Off-grid robustness check; not the trained $\psi$ or $c_{0.25}$.}
\label{tab:e4-by-regime}
\centering\small
\begin{tabular}{llrr}
\toprule
Arm & Regime & Prefix @0.30 & Ablation @0.40 \\
\midrule
No monitor & hacking & 0.945 & 0.938 \\
Probe s0 / s1 / s2 & hacking & 0.883 / 0.922 / 0.836 & 0.875 / 0.938 / 0.812 \\
SCL s1 / s2 & hacking & 0.008 / 0.016 & 0.008 / 0.016 \\
Cut s0 & hacking & 0.008 & 0.039 \\
SCL s0 & mixed & 0.023 & 0.000 \\
Cut s1 / s2 & mixed & 0.070 / 0.000 & 0.117 / 0.000 \\
\bottomrule
\end{tabular}
\end{table}

\subsection{Probe diagnostics}

The in-loop probe is read after an appended self-report turn, whereas the gate and two-reader diagnostic read the end of emitted code. Thus the gate AUROC does not measure the exact scorer optimized during training. Probe training logs give an in-loop score of $\sim10^{-9}$ from step 1 through 120.

\begin{table}[htbp]
\caption{Two-reader re-scoring with the frozen probe ($n=128$ per run). The reference reader is the baseline adapter; the policy reader is the run's own adapter. Reader means interleave across behavioral regimes, so the descriptive patterns test possible text or representation changes but do not identify control. Runs below hack share $0.5$ are not labeled.}
\label{tab:readers-multiseed}
\centering
\small
\begin{tabular}{llrrrl}
\toprule
Arm & Seed & Hack $d$ & $\bar q$ reference & $\bar q$ policy & Pattern \\
\midrule
No monitor & --- & 0.984 & 0.750 & 0.750 & neither \\
Probe & 0 & 0.953 & 0.324 & 0.580 & policy-reader drop \\
Probe & 1 & 0.969 & 0.508 & 0.682 & reference-reader drop \\
Probe & 2 & 0.961 & 0.582 & 0.638 & policy-reader drop \\
SCL & 0 & 0.156 & 0.245 & 0.540 & --- (mixed) \\
SCL & 1 & 0.938 & 0.689 & 0.782 & neither \\
SCL & 2 & 0.883 & 0.609 & 0.247 & policy-reader drop \\
Cut & 0 & 0.930 & 0.555 & 0.699 & reference-reader drop \\
Cut & 1 & 0.133 & 0.223 & 0.082 & --- (mixed) \\
Cut & 2 & 0.094 & 0.102 & 0.160 & --- (mixed) \\
\bottomrule
\end{tabular}
\end{table}

\subsection{Continuation budget}\label{app:continuation-budget}

Because forced continuations are capped at 64 tokens, a final program longer than the cap cannot be reproduced and would score zero by construction. Tokenizing the fenced code body with the Llama-3.1-8B tokenizer, medians are 9--31 tokens and the share exceeding the cap (including the two-token closer) is 0/128 in every no-monitor, Probe, and Cut seed-0 run, 1/127 in SCL seed 1, and at most 11/128 in any mixed run. Per-continuation cap-hit flags were not written to the endpoint log, so we bound the mechanism by program length rather than measuring truncation directly.

\begin{table}[htbp]
\caption{Per-run median program lengths under the Llama-3.1-8B tokenizer. All medians are well below the 64-token forced-continuation cap.}
\label{tab:continuation-budget}
\centering
\small
\begin{tabular}{lllr}
\toprule
Arm & Seed & Regime & Median code tokens \\
\midrule
No monitor & --- & hacking & 10 \\
Probe & 0 & hacking & 11 \\
Probe & 1 & hacking & 9 \\
Probe & 2 & hacking & 11 \\
SCL & 0 & mixed & 31 \\
SCL & 1 & hacking & 13 \\
SCL & 2 & hacking & 18 \\
Cut & 0 & hacking & 11 \\
Cut & 1 & mixed & 24 \\
Cut & 2 & mixed & 20 \\
\bottomrule
\end{tabular}
\end{table}

\end{document}